\documentclass[a4paper,conference]{ieeeconf}

\IEEEoverridecommandlockouts                              %

\usepackage{graphicx} %
\usepackage{amsmath} %
\usepackage{amssymb}  %
\usepackage{amsfonts}
\usepackage[T1]{fontenc}
\usepackage{fix-cm}
\usepackage[hidelinks]{hyperref}
\usepackage[caption=false]{subfig}
\usepackage{tabulary}
\usepackage{tabularx}
\usepackage{url}
\usepackage{booktabs}
\usepackage{multirow}
\usepackage{xfrac}
\usepackage{cite}
\usepackage{microtype}
\graphicspath{{./figures/}}

\newcommand{\citep}[1]{\cite{#1}}
\newcommand{\citet}[1]{\cite{#1}}

\newcommand{\work}{paper}

\newcommand{\RR}{\mathbb{R}}

\title{\huge Neural-Attention-Based Deep Learning Architectures for Modeling Traffic Dynamics on Lane Graphs}

\author{Matthew A. Wright, Simon F. G. Ehlers, and Roberto Horowitz%
\thanks{M.A.W. and R.H. are with the Mechanical Engineering Department, Partners for Advanced Transportation Technologies, and Berkeley DeepDrive, University of California, Berkeley, CA, USA
        {\tt\small \{mwright, horowitz\}@berkeley.edu}}%
\thanks{S.F.G.E. is with the Institute of Transport and Automation Technology, Leibniz Universit\"{a}t Hannover, Hannover, Germany (work done while visiting UC Berkeley)
        {\tt\small simon.ehlers@stud.uni-hannover.de}}%
}%
\begin{document}

\maketitle

\begin{abstract}
Deep neural networks can be powerful tools, but require careful application-specific design to ensure that the most informative relationships in the data are learnable.
In this paper, we apply deep neural networks to the nonlinear spatiotemporal physics problem of vehicle traffic dynamics.
We consider problems of estimating macroscopic quantities (e.g., the queue at an intersection) at a lane level.
First-principles modeling at the lane scale has been a challenge due to complexities in modeling social behaviors like lane changes, and those behaviors' resultant macro-scale effects.
Following domain knowledge that upstream/downstream lanes and neighboring lanes affect each others' traffic flows in distinct ways, we apply a form of neural attention that allows the neural network layers to aggregate information from different lanes in different manners.
Using a microscopic traffic simulator as a testbed, we obtain results showing that an attentional neural network model can use information from nearby lanes to improve predictions, and, that explicitly encoding the lane-to-lane relationship types significantly improves performance.
We also demonstrate the transfer of our learned neural network to a more complex road network, discuss how its performance degradation may be attributable to new traffic behaviors induced by increased topological complexity, and motivate learning dynamics models from many road network topologies.
\end{abstract}

\section{Introduction}
Well-designed deep neural networks (deep NNs) have exhibited a powerful ability to model complex nonlinear phenomena.
While deep NNs may be most well known for their uses in image processing and statistical language modeling applications \citep{lecun_deep_2015}, this modeling ability also makes them well-suited to predicting nonlinear physical phenomena like the dynamic behavior of road traffic \citep{lv_traffic_2015}.

In all deep NN applications, the particular neural architecture (i.e., the type of ``layers'' used) is of particular importance: it is desired to select an NN with a so-called ``inductive bias'' that admits learning the necessary element-to-element relationships in a generalizable form \citep{battaglia_relational_2018}.

In this work, we present results on using NNs to model traffic dynamics at a lane level.
We consider the road network as a directed graph, with the nodes being road lanes and various types of edges being different kinds of lane-to-lane relationships.
We use an NN model to estimate macroscopic traffic characteristics for these lanes (e.g., the number of vehicles occupying it at a given time).

The motivation to consider macroscopic dynamics for lane graphs (rather than the traditional road dynamics graph of ``links,'' with lanes lumped together \citep{treiber_traffic_2013}) is threefold.
First, lanes are more of an ``atomic unit'' of the road network:
while statistical models for individual lanes may freely be composed into one for the whole link, the reverse is not true.

Second, resolution of road features at a lane level is necessary for vehicle control and route planning applications.
An NN architecture that can take in per-lane data and output per-lane network predictions would be useful in these settings.

Third, a lane-level NN model can address a shortcoming of traditional macroscopic traffic dynamics modeling.
Macroscopic traffic models can describe the aggregate behavior of all vehicles on a road via a continuous partial differential equation (PDE) approximation \citep{treiber_traffic_2013,ferrara_freeway_2018}.
Zooming in to the lane level, however, requires modeling of more complex human behaviors (lane changes, for example, have proven difficult to model, both in predicting when drivers will make a maneuver and how it will affect the other vehicles on the road \citep{zheng_recent_2014}).
This gap in our first-principles understanding is a natural place to attempt to statistically learn a model rather than derive one.

In this \work, using a microscopic traffic simulator \citep{krajzewicz_recent_2012} as a testbed, we find that a lane-graph-based NN can achieve these goals.
We present results showing that a deep NN can outperform a PDE-model-based method on an intersection queue estimation problem.
The NN architecture leverages the lane graph, basing per-lane queue and occupancy estimates on upstream, downstream, and neighboring lanes' observations.
We show that encoding these different lane-to-lane relationships in the lane graph differently (which some earlier NN layers for graph-structured data do not do) is critical for this problem.
We also show that our NN architecture can permit the transfer of NN models between road networks, opening the door to cross-region data-driven intelligent transportation systems.
All code used in development of these results is available online at \texttt{\href{https://github.com/mawright/trafficgraphnn}{github.com/mawright/trafficgraphnn}}.

The remainder of this \work~is organized as follows.
Section \ref{sec:background} briefly reviews the idea of aggregate traffic models, NN operations for graph-structured data, and the specific application of deep NNs to road networks.
All three problems have a large body of work; we review only a few closely-related items here.
Section \ref{sec:layer} presents the new NN layer we use in our lane graph models.
Section \ref{sec:methods} presents our developed software framework and the estimation problem we consider as a test case.
Section \ref{sec:implementation} provides details for our simulation setup and NN methods.
Section \ref{sec:results} presents in detail the results we outlined in the previous paragraph, and section \ref{sec:conclusion} outlines next steps.

\section{Background}
\label{sec:background}
\subsection{Partial Differential Equation Models of Traffic}
\label{sec:pde}
Traffic flows at an aggregate level are often described by analogy to compressible fluid flows \citep{treiber_traffic_2013}.
Mathematically, 
this means a fluid-like partial differential equation (PDE).
The PDE model of traffic is usually credited to Lighthill and Whitham \citet{lighthill_kinematic_1955} and Richards \citep{richards_shock_1956}.

The PDE models of traffic, while crude approximations of traffic's dynamics, have been invaluable tools in traffic engineering, both for forecasting/infrastructure planning and for real-time traffic control \citep{ferrara_freeway_2018}.

In this \work, we make use of a particular method based on the PDE model of traffic \citep{liu_real-time_2009}.
The method of \citet{liu_real-time_2009} is a technique for estimating the number of vehicles queued at a red light, in situations where this value is not directly observable.
More specifically, it is often the case that an intersection is instrumented only with inductive loop detectors that can measure when a vehicle is occupying a specific location, and if the queue at an intersection extends past the most-upstream detector, it is unobservable.

In \citet{liu_real-time_2009}, it is proposed to apply the PDE approximation of traffic to estimate this unobserved queue.
After the light turns green and traffic begins to move, \citet{liu_real-time_2009} propose to identify when the shockwaves predicted by the PDE model of traffic passes by these fixed detectors, then to apply a method-of-characteristics analysis to track the shock trajectories in space and time and finally back out the upstream PDE boundary location (i.e., the initial length of the queue).
This calculation is performed on each lane individually to obtain a \emph{per-lane, per-cycle} maximum queue estimate.
There are obvious inaccuracies to applying a continuous PDE analysis at a scale where the continuum approximation is not valid, but nevertheless, the method represents an integration of domain knowledge to obtain an estimate where there was none before.

In this \work, we make use of \citet{liu_real-time_2009}'s method both as a model-based baseline, and as an input to our NN models.
For more details on our implementation, see \citet{ehlers_traffic_2019}.

\subsection{Neural network operations for graph data}
Deep NNs have had particular success in domains with regularly-structured data, such as images and text.
In those domains, prior knowledge that the data structure encodes important information (e.g., that a pixel's meaning can be better understood by examining it in combination with nearby pixels than with far-away pixels) enable better statistical learning.
The term \emph{geometric deep learning} \citep{bronstein_geometric_2017} has been used to describe efforts to generalize these lessons to structured data on more general topological domains, such as graphs.

Many early works on deep NNs for graph-structured data consider learning features in the graph's spectral domain (i.e., in the eigenbases of matrices associated with the graph) \citep{kipf_semi-supervised_2017,defferrard_convolutional_2016}.
Another line of work (~\citep{hamilton_inductive_2017,velickovic_graph_2018}, etc.) attempts to make inferences in the ``graph domain,'' i.e., by relating nodes to their neighbors (c.f. the difference between how modern convolutional NNs' operations are carried out in the pixel/time domain, whereas in classical signal processing, convolutions are often performed in the frequency domain).

One item to note is that, for features learned in the spectral domain of a graph, those features are tied to the particular graph spectrum.
The transferability of spectral graph NNs is unclear \citep{bronstein_geometric_2017}.
The neighborhood-based approaches, on the other hand, do not have this constraint and can be freely transferred (but, of course, have the drawback that a node's ``receptive field'' is limited).
In this \work, we are explicitly interested in transferring between lane graphs, and so our proposed models fit in the neighborhood-based approach (we also present comparisons against a spectral-type approach \citep{kipf_semi-supervised_2017} in our detailed results).

More broadly, authors have recently argued that AI systems need to be allowed to learn to reason about related but distinct entities to truly mimic humans' learning capability \citep{battaglia_relational_2018}.
The existence of distinct entities with information tied to their relationships to one another is often rendered mathematically as graphs, and inferring meaning about their relationships as classic message passing; designing NN architectures that can approximate this tractably is a topic of much discussion \citep{battaglia_relational_2018}.

Our particular NN is based largely on a neural architecture termed \emph{attention} (\citet{bahdanau_neural_2015,vaswani_attention_2017}, etc.).
This technique and our application of it are described in section \ref{sec:layer}.

\subsection{Neural networks for traffic dynamics}
We are of course not the first to use deep NNs for traffic flows.
The difficulty of first-principles modeling of traffic flow, along with the availability of massive amounts of observational data, make it a model problem for machine learning methods \citep{lv_traffic_2015}.
Here, we briefly discuss two recent works that make use of graph-type NN operations.

\citet{li_diffusion_2018} and \citet{cui_traffic_2018} both propose graph NN layers that learn spatiotemporal relationships in traffic networks.
\citet{li_diffusion_2018} propose a ``diffusion convolutional'' operation with learned coefficients, used to model traffic movements as a random walk along a graph.
\citet{cui_traffic_2018} propose a ``traffic graph convolutional'' operation that incorporates traffic-theoretic information by clipping adjacency matrices based on whether traffic can feasibly move from one location to another in a time interval, based on the road distance and a prescribed freeflow speed.

In both papers, this graph operation forms the internals of a recurrent neural network (RNN) layer that is iteratively applied to a traffic timeseries to learn an autoregressive model.
Their methods are applied to freeway-network-scale traffic measurements in two major U.S. metro areas: Los Angeles in \citet{li_diffusion_2018} and Seattle in \citet{cui_traffic_2018}.

In comparison, while in the present \work~we do not attempt a joint learning of spatial and temporal relationships via a spatial RNN layer (see section \ref{sec:architecture} for our NN architecture), we operate at a finer scale (lane level) and explicitly consider the graph-to-graph transfer problem.
We also consider the impact of having multiple edge types.
This last point is of importance at a lane graph level, since the macroscopic meaning of lane-to-lane relationships are distinctly different for adjacent lanes (i.e, lane changes) and upstream/downstream lanes.

\section{Multi-edge-type Attentional Neural Network Layers}
\label{sec:layer}
Neural attention \citep{bahdanau_neural_2015} is a powerful and widely-applicable neural architecture.
Neural attention is often described as letting the NN ``attend'' or ``pay attention to'' certain elements of the input data \citep{vaswani_attention_2017}.
This means that, in a multilayer NN, an early layer is able to send to later layers the most relevant information for downstream tasks.

Technically, we define an NN layer for graph data as accepting two items, 1) a graph $(V, E)$ with node set $V$ and edge set $E$ and 2) $\{x_i \in \RR^N : i \in V\}$, a data vector for every node.
In this work, we restrict our focus to layers where the domain of the output data is the same graph $(V, E)$.
That is, the output of the NN layer is $\{h_i \in \RR^H : i \in V\}$, a per-node featurization of the input data.

In this \work, we assume that our graph is a directed graph, and also allow the graph to be multidimensional, with each edge also having one of a finite number of edge dimensions $d \in \{1,\dots,D\}$.
We denote the collection of $d$-dimension edges as $E^d$.
In our application, the edge dimensions correspond to different lane-to-lane topological relationships, e.g., upstream, downstream, adjacent, self, etc.

For computational purposes, define a fixed ordering for the nodes $\{1, \dots, n\}, n = |V|$ and the corresponding per-edge-type adjacency matrices $A_d$, where the $i,j$th entry of $A_d$ is 1 iff $(i,j) \in E^d$, and 0 otherwise.

The basic outline of a standard attentional layer is to learn two tasks in parallel: 1) a learned featurization of each node's data $f_i = f(x_i)$ and 2) an attention scoring $a:\RR^N \times \RR^N \to \RR$ that quantifies the relative importance of the edge $(i,j)$.
Then, the attention scores are used to compute, for each node, a weighted average of the featurized data of all nodes (or some subset of them).

In particular, we adapt the ``Graph Attention Layer'' of \citet{velickovic_graph_2018} with a new generalization to multidimensional graphs.
We perform our per-node embedding via a learned weight matrix $W \in \RR^{F \times N}$, $f_i = W x_i$.
Then, we compute the attention scoring with a single (scalar output) fully-connected neural network layer, one for each edge type:
\begin{align}
    \begin{split}
    a_{i,j}^d &= \sigma\left(\left[
        \left(W_i^d\right)^T \; \left(W_j^d\right)^T\right]
        \left[\left(f_i\right)^T \; \left(f_j\right)^T\right]^T\right) \\
        &= \sigma\left(
            \left(W_i^d\right)^T f_i + \left(W_j^d\right)^T f_j
        \right)
    \label{eq:attn_scores}
    \end{split}
\end{align}
where $W_i^d, W_j^d \in \RR^{F}$ are learned weights and $\sigma(\cdot)$ denotes some elementwise NN nonlinearity (in this \work, following \citet{velickovic_graph_2018}, we use for $\sigma(\cdot)$ the LeakyReLU function with slope parameter 0.2).

The attention scores $a_{i,j}^d$ are then normalized via a softmax function,
\begin{equation}
    \alpha_{i,j}^d = \frac{\textnormal{exp} (a_{i,j}^d)}{\sum_{\{k: (i,k) \in E^d\}} \textnormal{exp} (a_{i,k}^d)}.
    \label{eq:softmax}
\end{equation}

The per-node, per-edge-type layer outputs are computed as a weighted average of $f_j$ for nodes $j$ to which $i$ is $d$-related:
\begin{equation}
    h_i^d = \Bigg(\sum_{\{k: (i,k) \in E^d\}} \alpha_{i,k}^d \, f_k\Bigg) + b^d
    \label{eq:avg}
\end{equation}
where $b^d$ is a learned bias vector.

Finally, the per-edge-type weighted-average feature vectors are concatenated together,
\begin{equation}
    h_i = \left[
        \left( h_i^1 \right)^T \; \cdots \;
        \left( h_i^D \right)^T
    \right]^T
    \label{eq:edgetype_concat}
\end{equation}
and this vector $h_i$ becomes the layer output for node $i$.

We make use of \emph{multi-head attention} \citep{vaswani_attention_2017}, where the above calculations are all performed several times, in parallel.
The idea is that different attention ``heads'' are able to learn different concepts.
In a multi-head layer, every head's $h_i$ \eqref{eq:edgetype_concat} is concatenated together to become the final layer output.
Subsequent layers then combine information across edge types and attention heads.
Like \citet{vaswani_attention_2017}, we use fully-connected layers for this purpose.

\section{Methods}
\label{sec:methods}
\subsection{Software framework}
We wrote Python code to set up traffic simulations, extract and preprocess data, and train the NNs.
We used the microscopic traffic simulator SUMO (Simulation of Urban MObility) \citep{krajzewicz_recent_2012}.
Our NN layers use the Keras \citep{chollet2015keras} ``Layer'' standard in the interest of making them generally applicable.

\subsection{A model problem for spatiotemporal deep learning on lane graphs: traffic queue prediction}
Our test case problem in this \work~is the estimation of the number of vehicles on a lane preceding a traffic signal, using readings from traditional fixed loop detectors.
Accurate real-time vehicle counts at a traffic signal are necessary for effective signal control \citep{liu_real-time_2009,kurzhanskiy_traffic_2015}, but, as discussed in section \ref{sec:pde} where we describe a model-based estimation method, these counts are in general unobservable from raw data alone.

We define two output quantities for our NN to predict: the maximum queue length at any point in a lane's red-green cycle, and the total count of vehicles on the lane at every timestep (the second quantity will be always no smaller than the first since it includes vehicles in transit who have not reached the signal queue yet).

\section{Implementation Details}
\label{sec:implementation}
\subsection{Simulation Details}
\label{sec:sim_details}
\begin{figure*}[t]
    \centering
    \subfloat[]{\includegraphics[width=2.3in]{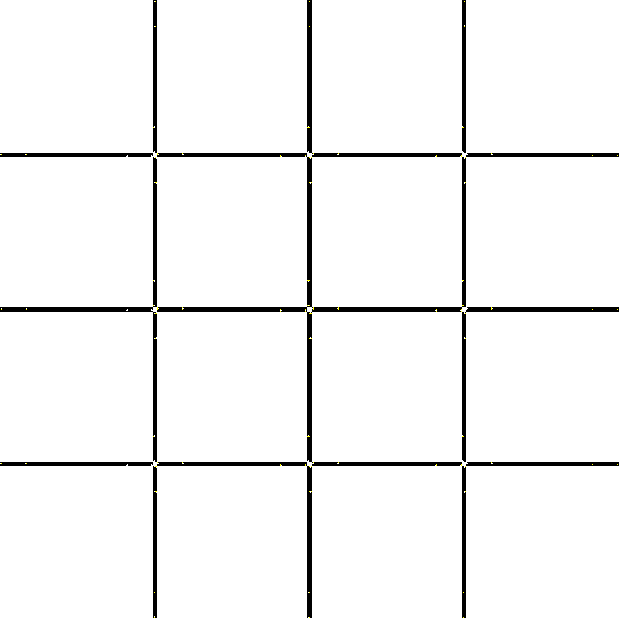}
    \label{fig:grid_far}} \hfil
    \subfloat[]{\includegraphics[width=2.5in]{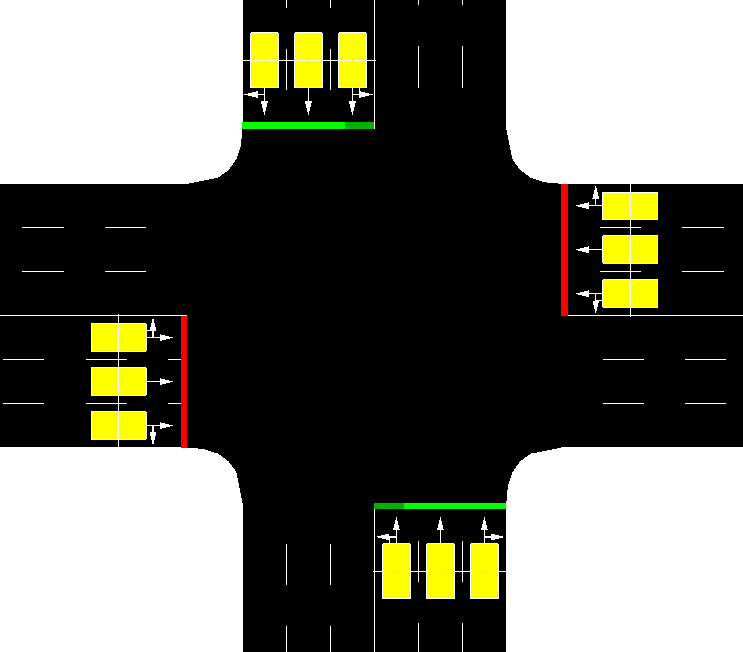}
    \label{fig:grid_zoom}}
    \caption{Training road network. (a) A 3x3 grid of three-lane roads, with attaching perimeter roads. Each road is 750 m long. (b) Zoomed view of one intersection. The yellow boxes represent stopbar vehicle detectors.}
    \vspace{-10pt}
    \label{fig:grid}
\end{figure*} 
We generated 100 SUMO simulations on the three-by-three grid network shown in Figure \ref{fig:grid}.
For each simulation, we generated a sequence of random vehicle trips through the road network using SUMO's \texttt{randomTrips} tool.
Each vehicle entered the network on a randomly-selected entrance road (one of the roads whose upstream boundary is the edge of the road network) and was given a goal of a randomly-selected exit road (one of the roads whose downstream boundary is the edge of the network).
A new vehicle entered the road in this manner every 0.4 s, for 3600 s.
Each vehicle had a maximum speed given by multiplying SUMO's default maximum speed by a randomly-drawn multiplicative factor drawn from a Gaussian distribution with mean 1 and standard deviation 0.1, clipped at 0.2 and 2.
We enabled SUMO's ``rerouting'' option so vehicles could reroute themselves around congestion during the simulation, and disabled its default ``teleport'' option.
All other simulation parameters were left at their defaults.

On each 750-m lane, we placed two simulated loop detectors: one at the stopbar (visible in Fig \ref{fig:grid_zoom}) and one 125 m upstream.
Each detector recorded data every 1 s.
These data are occupancy (a value in $[0,1]$ reporting the portion of the 1-s period that a vehicle occupied the loop detector) and the speed of those vehicles (if any).

The input data for lane $i$ at timestep $t$, $x_i^t \in \RR^6$ was a vector with elements 1) the stopbar detector occupancy, 2) the stopbar detector speed, 3) the upstream detector occupancy, 4) the upstream detector speed, 5) a 0-1 indicator value that is 1 if the lane's traffic light is green and 0 if it is red, and 6) the PDE-model-based \citet{liu_real-time_2009} estimate of the queue.

We note that, when the traffic queue is not detected to have extended past the upstream detector, \citet{liu_real-time_2009} notes that the per-lane queue can be estimated via an input-output accounting.
In practice, we found \citep{ehlers_traffic_2019} that this was especially sensitive to lane changes.
Therefore, when a per-lane input-output difference resulted in a negative number of cars, we use a heuristic where we split this deficit evenly among the lanes that accounted for a surplus of cars (in reality, they had lane-changed out of the lane with a surplus and counted against the lane who saw it exit without entering).

The output vector $y_i^t \in \RR^2$ has elements 1) the maximum queue length, in vehicles, of the lane during a red-green cycle, and 2) the count of the total number of vehicles on the lane during that timestep.
These values were recorded via SUMO's ``lane area'' detectors and are given by the record fields \texttt{maxJamLengthInVehicles} and \texttt{nVehSeen}.

The queue length and the PDE-based queue length estimate are defined \emph{per cycle} rather than \emph{per timestep}.
The PDE-based estimate \citep{liu_real-time_2009} gives a per-lane estimate of the time of the maximum queue (based on the PDE characteristic curves), while the input-output relationship (used when the queue length does not extend past the upstream detector) can give an instantaneous per-lane estimate when the light changes from green to red.
All other timesteps had this feature filled with the dummy value -1.

\subsection{Our neural network architecture}
\label{sec:architecture}
In the NN literature, the problem setting where both the input and output domains are sequentially structured (i.e., timeseries data) is called the ``sequence-to-sequence'' problem \citep{sutskever_sequence_2014,neubig_neural_2017}.
The canonical NN architecture for this problem is a so-called ``encoder-decoder'' \citep{sutskever_sequence_2014} approach.

One approach to modeling spatiotemporal data such as videos \citep{venugopalan_sequence_2015} is to break the encoder and/or decoder into two parts.
First, an encoder for the spatial information (i.e., a convolutional NN stack or graph-type layers) operates on each timestep independently.
Then, these per-timestep encodings are passed into an NN architecture meant to learn temporal correlations, such as an RNN layer stack.
Our NN model used in this \work~follows this paradigm.

We take in a set of per-lane input timeseries $\{x_i^t \in \RR^N \}_{t=1}^T$ for lanes $i \in \{1,\dots,n\}$.
We pack each timestep's data into a data matrix for the entire graph, $X^t \in \RR^{n \times N}$
We pass each data matrix into a graph encoder stack made up of several layers, each layer consisting of two sublayers: first, one of our multi-edge-type attention layers; second, a fully-connected layer.
After each sublayer, we apply a layer normalization \citep{ba_layer_2016} operation with learned scaling and shift parameters.
The idea of a layer with an attentional sublayer followed by a fully-connected sublayer is inspired by the Transformer architecture of \citet{vaswani_attention_2017}, though our fully-connected sublayer is simpler and we omit the residual connections.
In this \work, our graph encoder stack has two of these attention-fully connected layers, with ReLUs in between each sublayer.

The output of this graph encoding layer is a matrix of size $n \times F$, where $F$ is the feature dimension of the last fully-connected layer.
We have $\texttt{batch\_size} \times T$ of these matrices, one for each timestep and element of our sample batch.
The objective is that each node's $F$-long feature vector has aggregated information from other nodes' data.

Our next step is to learn the temporal relationships.
We pass each of the $\texttt{batch\_size} \times n$ timeseries into an RNN independently.
In this work, we use a two-layer RNN, using the gated recurrent unit (GRU) architecture.
The first RNN layer is meant to encode the sequence forward in time.
The second RNN layer uses an attentional decoding mechanism based on \citet{bahdanau_neural_2015}, with a modification that we do not let the RNN attend to timesteps in the future, using a masking technique similar to \citet{vaswani_attention_2017,velickovic_graph_2018} and our graph encoder.
We use ReLU nonlinearities at the output of each RNN layer as well.

The final layer in our NN is a fully-connected layer that maps the output of the second RNN layer to our output domain $\{y_i^t \in \RR^M \}$.
Since, in this case, our outputs (the queue length and the lane occupancy) are strictly nonnegative, we use a ReLU nonlinearity at the NN output as well, to restrict our NN's image to nonnegative numbers.

\subsection{Training details}
The 100 simulations were split into an 80-simulation training set, a 10-simulation validation set, and a 10-simulation test set.
We trained against the Huber loss.
The NNs were trained with the Adam optimization algorithm \citep{kingma_adam_2015}, with its parameters left at their Keras defaults.
We annealed the Adam learning rate by multiplying it by a factor of 0.1 when the Huber loss on the validation set did not decrease over 10 epochs.
We used early stopping, ending training when the validation Huber loss did not decrease for 20 epochs.

The data for each simulation were chopped into periods of 30-second averages.
For occupancy, speed, and the green indicator in $x_i$ and vehicles on the lane in $y_i$, we took the mean over the period, and for the defined-per-cycle values of maximum queue length in $y_i$ and PDE-based max queue estimate in $x_i$, we took the max.
This 30-second averaging of detector data is prevalent in traffic engineering \citep{jia_pems_2001}, despite recent awareness that it averages out important information \citep{coifman_empirical_2015}.
This choice then mimics a lossy downsampling that is endemic to commonly-available data for practitioners.
Another reason for doing this averaging is that we found that our NNs performed poorly on 1-second data. This is not too surprising; the 1-second data is very sparse, and it is known that RNNs have difficulty learning long-term dependencies \citep{trinh_learning_2018} despite many developments (including gated RNN architectures like the GRU) to tackle this problem.

For timesteps where the maximum queue length in $y_i$ is undefined (see the last paragraph of section \ref{sec:sim_details}), we did not use the NN's output for that output feature and no gradient was taken.

\section{Results and Discussions}
\label{sec:results}
We trained a variety of NNs with different combinations of encoded edge types in our layer from section \ref{sec:layer}.
Table \ref{tab:results} gives quantitative results in terms of estimation mean absolute error (MAE) on the 10-simulation held out test set.

In the table, the leftmost column describes the adjacency matrix structure included in our graph encoding layers described in section \ref{sec:layer}.
The row for ``$A=I$'' indicates that no lane-to-lane adjacency information was encoded; instead, an identity matrix was passed in.
This effectively means that the lane was only able to ``attend'' to itself, i.e., the graph encoding layer acted functionally identically to a fully-connected layer.
Further entries in the first block of rows indicate that we add more adjacency matrices to encode more structure.
``$+ A_{\textnormal{downstream}}$'' means that we also include an adjacency matrix to denote downstream lane-to-lane connections, ``$+ A_{\textnormal{upstream}}$'' adds the reverse relation (i.e., the transpose of $A_\textnormal{{downstream}}$), and $+ A_{\textnormal{neighbors}}$ means we added a matrix that indicates whether lanes were neighbors (i.e., they were on the same road).
These additional adjacency matrices are cumulative: $+A_{\textnormal{upstream}}$ also has the downstream adjacency dimension, and $+ A_{\textnormal{neighbors}}$ has all four.

The rows ``$\{I, A_\textnormal{upstream}\}$'' and ``$\{I, A_\textnormal{neighbors}\}$'' denote NNs with only the two noted adjacency matrices (as opposed to the cumulative adding of edge types in the previous rows).

All NNs described in Table \ref{tab:results} used four attention heads in the graph encoding sublayers.
Information regarding parameter counts for each NN is given in the Appendix.

Next, we highlight several findings.

\setlength{\tabcolsep}{5pt}

\begin{table}
    \begin{center}
    \caption{PDE- and NN-based estimation results}
    \label{tab:results}
    \begin{tabular}{l r r}
        \toprule
        Configuration & \multicolumn{2}{c}{Mean Absolute Error (vehicles)} \\
        \cmidrule(lr){2-3}
        & \multicolumn{1}{c}{Cycle Queue Length} & \multicolumn{1}{c}{Lane Occupancy} \\ \midrule
        $A=I$ & 1.04 $\pm$ 0.004 & 1.50 $\pm$ 0.003 \\
        +$A_\textnormal{downstream}$ & 1.06 $\pm$ 0.01 & 1.47 $\pm$ 0.01 \\
        +$A_\textnormal{upstream}$ & 1.06 $\pm$ 0.01 & 1.37 $\pm$ 0.01 \\
        +$A_\textnormal{neighbors}$ & 0.96 $\pm$ 0.01 & 1.24 $\pm$ 0.01 \\
        \midrule
        $\{I, A_\textnormal{upstream}\}$ & 1.07 $\pm$ 0.004 & 1.40 $\pm$ 0.01 \\
        $\{I, A_\textnormal{neighbors}\}$ & 0.97 $\pm$ 0.01 & 1.40 $\pm$ 0.01 \\
        \midrule
        PDE-based estimate \citep{liu_real-time_2009} & 5.36 & - \\
        \bottomrule
    \end{tabular}
    \end{center}
    \footnotesize{Values shown are mean abs. error $\pm$ one std. dev. over five random seeds.}
    \vspace{-10pt}
\end{table}

\subsection{Significantly Outperforming the PDE-Based Estimate}
\begin{figure}[t]
    \centering
    \includegraphics[width=20pc]{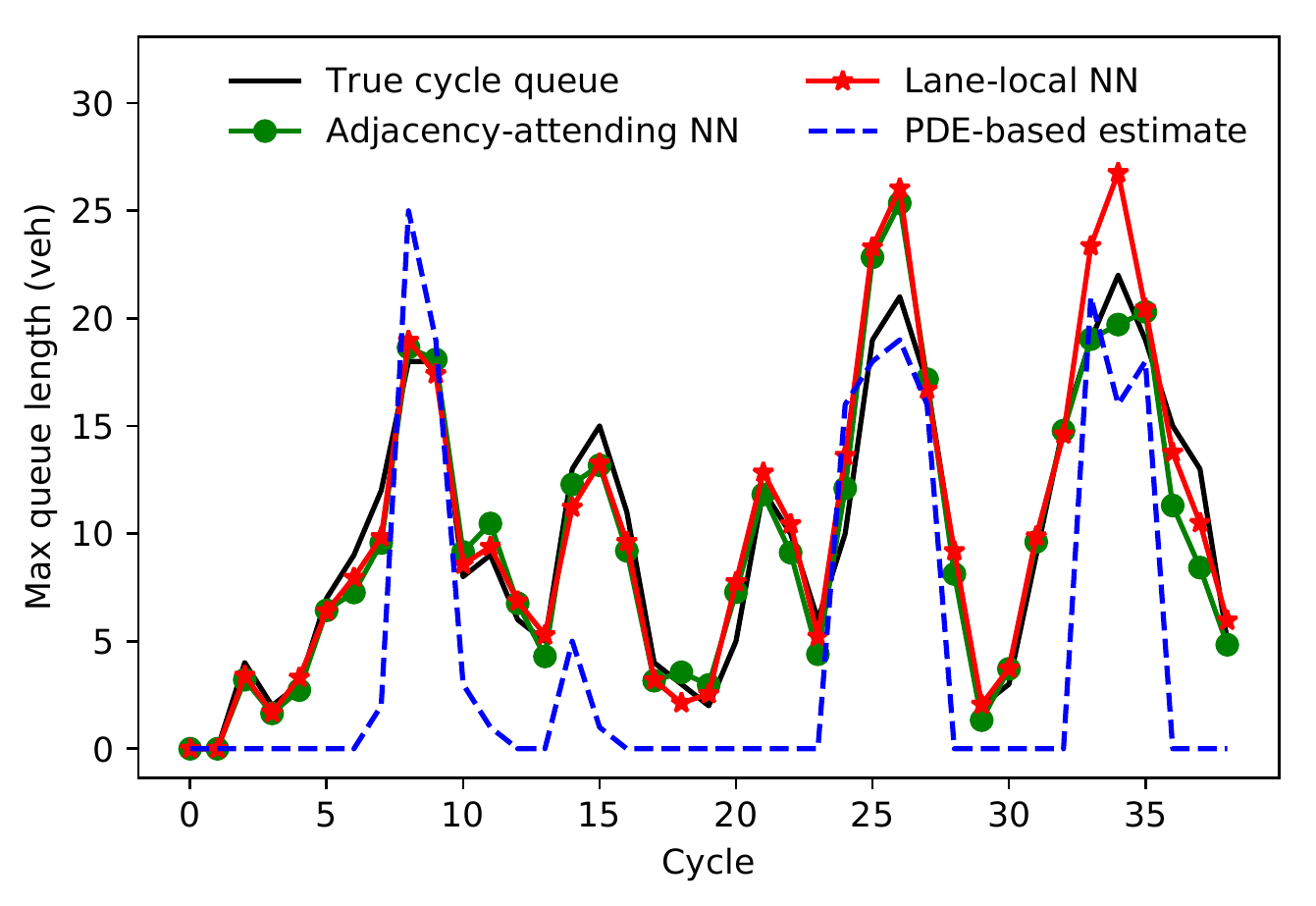}
    \vspace{-5pt}
    \caption{Example per-cycle queue estimates for a right-turn lane. The adjacency-attending NN is the one with all four edge-type encodings in Table \ref{tab:results} and the lane-local NN is the $A=I$ one.
    Note that the NN that can attend to adjacent lanes is able to predict the queue near cycle 35 more accurately than the lane-local one.
    When the queue does not back up enough to generate a shockwave to observe, the PDE-based method falls back to an input-output equation \cite{liu_real-time_2009}, which performs poorly in the presence of lane changing \cite{ehlers_traffic_2019}.}
    \label{fig:liu_vs_nn}
    \vspace{-10pt}
\end{figure}

We found that our lane-graph NN models significantly outperform the method of \citet{liu_real-time_2009} in estimating the lane queues.
The mean absolute error value of 5.36 vehicles is much higher than even our worst-performing NN.
Figure \ref{fig:liu_vs_nn} shows an example of the per-cycle lane queue estimates for one right-turn lane in the grid network.

Note that \citet{liu_real-time_2009}'s method was proposed to do lanewise queue estimation, but, as we mentioned earlier and discuss in more detail in \citet{ehlers_traffic_2019}, we found that it performs poorly in lanewise estimation in simulations with lane changing, like the ones analyzed in this paper.

\subsection{Benefits of Attending to Adjacent Lanes}
\label{sec:subsets}
We see in Table \ref{tab:results} that assimilating additional lanes' information via our multi-edge-type attentional encodings provides benefits to estimation.
This is particularly notable in the occupancy estimation problem, where each additional edge type added gives a performance boost.

Interestingly, the cycle maximum queue length estimation performance seems to do worse when the downstream and upstream lanes' information are attended to.
Under a two-sample $t$-test (unequal variances), the performance loss for both cases (adding the downstream lanes and adding both the downstream and upstream lanes) is statistically significant under most traditional levels ($p\approx0.005$ and $p\approx 0.0005$, respectively) (the difference in estimation error between the attending downstream case and the attending downstream and upstream case is not statistically significant, $p\approx0.41$).
The fact that performance actually degrades instead of holding steady likely indicates overfitting.

Adding the neighboring lanes' information gave a clear performance boost in the max queue length problem, indicating that attending to those lanes is useful.
Note also that while the configuration of only attending to neighboring lanes (the row $\{I, A_\textnormal{neighbors}\}$) achieves about the same performance on queue estimation as the NN with all four adjacency matrices, it cannot perform as well on the lane occupancy problem.
So, attending to \emph{all} types of adjacent lanes is beneficial for this other, harder, problem.

\begin{table}[t]
    \begin{center}
    \caption{Results for sections \ref{sec:ablation} and \ref{subsec:gcn}}
    \label{tab:ablation}
    \begin{tabular}{l r r}
        \toprule
        Variation from Base Model & \multicolumn{2}{c}{Mean Absolute Error (vehicles)} \\
        \cmidrule(lr){2-3}
        & \multicolumn{1}{c}{Cycle Queue Length} & \multicolumn{1}{c}{Lane Occupancy} \\ \midrule
        (approx.) $\sfrac{1}{4}$-size attn. dim. & 0.98 $\pm$ 0.01 & 1.27 $\pm$ 0.01 \\
        Flattened adjacency matrix & 1.22 $\pm$ 0.03 & 1.58 $\pm$ 0.03 \\
        One (4x-size) attention head & 0.96 $\pm$ 0.01 & 1.24 $\pm$ 0.02 \\
        \midrule
        GCN, $\{I, A_\textnormal{up/downsteam}\}$ & 1.81 $\pm$ 0.06 & 2.20 $\pm$ 0.05 \\
        GCN, $\{I, A_\textnormal{neighbors}\}$ & 1.87 $\pm$ 0.04 & 2.54 $\pm$ 0.04 \\
        GCN, all adj. matrices $A$ & 1.49 $\pm$ 0.03 & 1.85 $\pm$ 0.03 \\
        \bottomrule
    \end{tabular}
    \end{center}
    \vspace{-3pt}
    \footnotesize{``Base model'' refers to the model with all four adjacency matrices (+$A_\textnormal{neighbors}$ in Table \ref{tab:results}).
    GCN = Graph Convolutional Network \citep{kipf_semi-supervised_2017}.
    Values shown are mean abs. error $\pm$ one standard deviation over five random seeds.}
    \vspace{-12pt}
\end{table}

\subsection{Ablation Studies}
\label{sec:ablation}
We investigated the contribution of individual components of our NNs by selectively removing them.
The results are shown in the first set of rows in Table \ref{tab:ablation} and discussed here.

\subsubsection{The advantage of four-way edge type encoding is not simply due to an increased number of parameters}
As discussed in section \ref{sec:layer}, in this work we augmented our graph encoding layers with extra adjacency matrices defining extra lane-to-lane relations by concatenating $h_i^d$'s together \eqref{eq:edgetype_concat}.
This means that the output dimension of the graph encoding layer increases linearly with $D$, and thus the fully-connected sublayer immediately downstream has a weight matrix whose number of rows also increases linearly with $D$.
As can be seen in the Appendix, this can lead to a large increase in the total NN parameter count.

To verify that the increased performance of our extra edge-type encodings are not just due to increased parameter counts of downstream layers, we trained an NN with a (roughly) quarter-sized attention dimension $F$ (it was not exactly $\sfrac{1}{4}$ size, but was chosen to bring the parameter count close to the $A=I$ case, as shown in Table \ref{tab:paramcounts}).
Its performance is shown in the first row of Table \ref{tab:ablation}.
We see there is a slight performance drop from the full-size base model, but we still perform substantially better than the other models with fewer adjacency dimensions.
This suggests the high performance of the four-dimensional edge-type NN is indeed due to being able to attend to other lanes.

\subsubsection{Flattening out edge type information is very costly}
\label{sec:flattening}
To study whether encoding different lane relationships in different ways (i.e., letting the model attend to upstream lanes in different ways than to neighboring lanes), we trained a model where we ``flattened out'' the adjacency dimensions.
That is, here we used a single-dimension ($D=1$) adjacency matrix where the $i,j$ entry being 1 meant that lane $i$ and $j$ were related by any of the lane-to-lane relationships (self, downstream, upstream, neighboring).

The second row in Table \ref{tab:ablation} shows this model's performance.
We see this model actually performed much worse, even worse in fact than the naive, non-adjacent-attending model ($A=I$ in Table \ref{tab:results}).
This suggests that attending to distinct relation types differently is critical, and attempting to treat them the same can lead to model confusion.

\subsubsection{Is multi-head attention not needed when domain knowledge gives information of all edge-type relationships?}
The third row in Table \ref{tab:ablation} shows the results when we use only one attention head.
In this case, the attention head is 4x larger than the base model's four attention heads, meaning that the output dimension of this sublayer is the same.
Quite interestingly, we saw no significant performance drop in our application, whereas \citet{vaswani_attention_2017}, when applying attention to a language translation task with no multi-edge-type encodings, saw a clear performance degradation with a single-headed configuration.
We hypothesize this may be because our multi-edge-type encoding captures the benefits of learning multiple types of relationships that multi-head attention captures.

\subsection{Comparing multi-edge-type attention to the Graph Convolutional Network layer}
\label{subsec:gcn}
The ``Graph Convolutional Network'' (GCN) layer of \citet{kipf_semi-supervised_2017} is a popular spectral-type graph NN layer.
The second group of rows in Table \ref{tab:ablation} presents results where replace our multi-edge-type attentional sublayers in our NN architecute with 256-hidden-unit GCN layers (using the Keras GCN implementation published by the authors, and using \citet{kipf_semi-supervised_2017}'s adjacency matrix ``renormalization trick''), with subsets of adjacency matrices (as in section \ref{sec:subsets}).

The GCN layer considers only one adjacency matrix and only undirected graphs.
This means that to apply the GCN layer to our setting, we need to both flatten the adjacency matrices into one (see section \ref{sec:flattening}) and drop all directionality information.
This means, for instance, that we cannot encode downstream and upstream relations separately for the GCN.
These topological assumptions not being valid for the nonlinear dynamics encountered on the road network may explain the GCN layer's poorer performance here.

\subsection{Graph-to-Graph Transfer Learning: First Results}
An important problem in geometric deep learning is the graph-to-graph transfer problem \citep{bronstein_geometric_2017}.
In the classic spectral-based graph NN layers, the layer features are computed in the graph's spectral domain.
This means that the applicability of these layers across graph topologies is undefined.

The local graph layers, on the other hand, since they rely only on local information, do not have this problem.
Further, the particular learned attention scoring construction of attentional graph layers \eqref{eq:attn_scores}-\eqref{eq:avg} means that different-sized $d$-neighborhoods can be used with the same layer.

\begin{figure}[t]
    \centering
    \includegraphics[width=12pc]{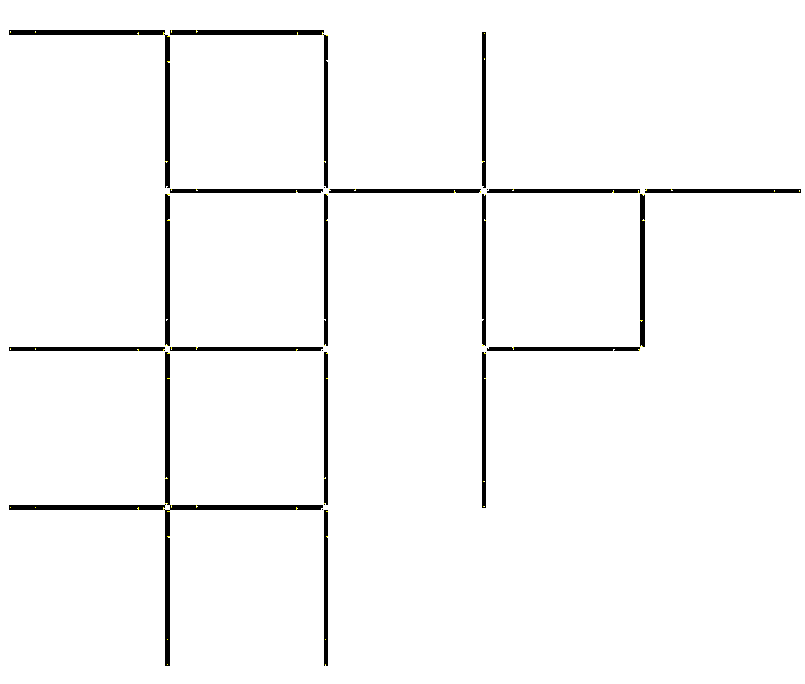}
    \caption{Random road network used in graph-to-graph transfer trial.}
    \label{fig:rand}
\end{figure}

Figure \ref{fig:rand} shows a randomly-structured grid-like network with 750-m-long roads (the same length as the grid network), generated with SUMO's \texttt{NETGENERATE} function.
We evaluated our NNs on 10 one-hour-long simulations on it.

\begin{figure}[t]
    \centering
    \vspace{-4pt}
    \includegraphics[width=20pc]{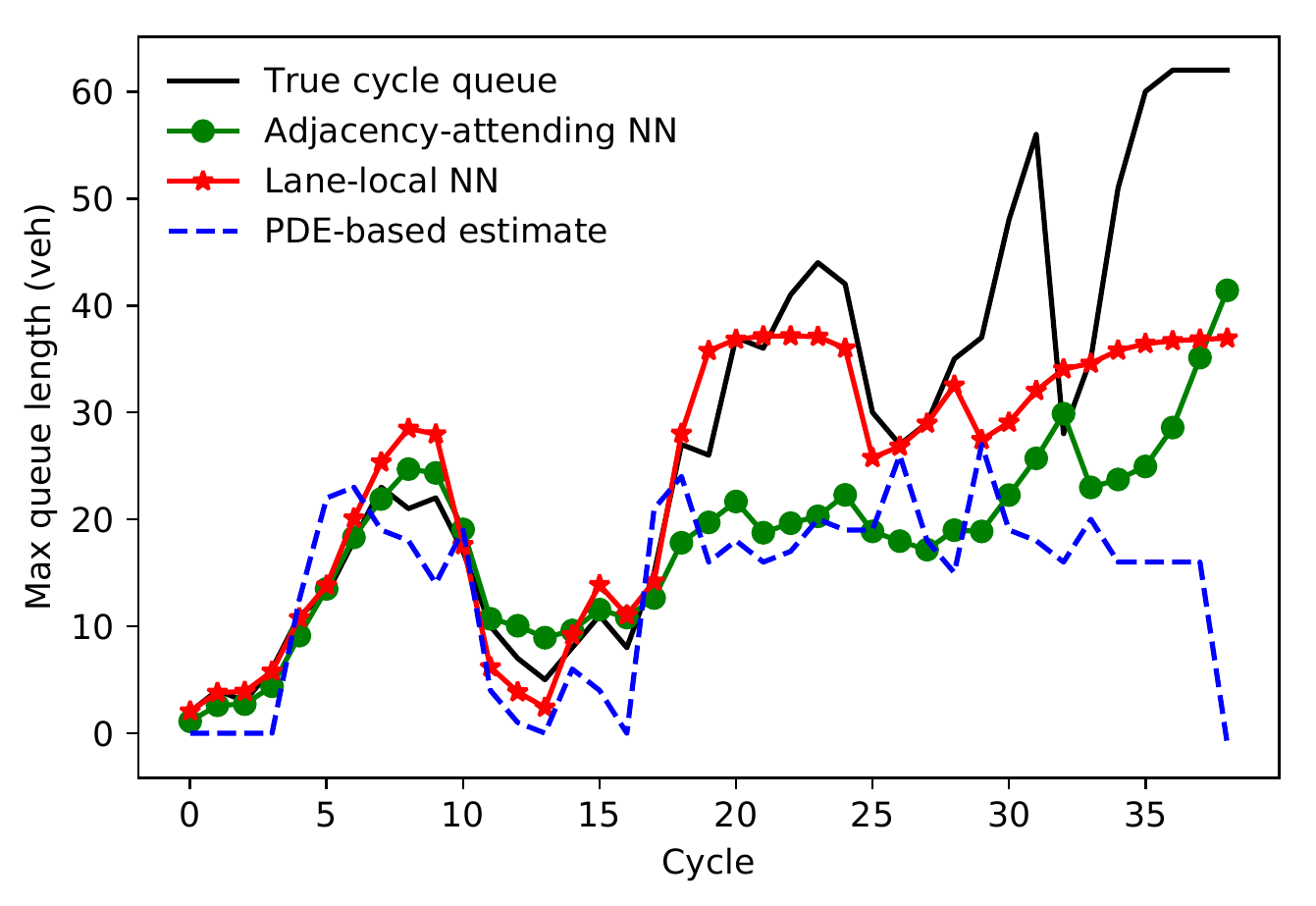}
    \caption{Example performance of the transferred NN from the grid network to the random network. The queue on a right-turn lane is shown.}
    \label{fig:g2g}
    \vspace{-12pt}
\end{figure}
 
\begin{table}[t]
    \begin{center}
    \caption{Graph-to-graph transfer results for two exemplar NNs}
    \begin{tabular}{l r r}
        \toprule
        Configuration & \multicolumn{2}{c}{Mean Absolute Error (vehicles)} \\
        \cmidrule(lr){2-3}
        & Cycle Queue Length & Lane Occupancy \\
        \midrule
        $A=I$ & 5.77 $\pm$ 0.10 & 7.47 $\pm$ 0.14 \\
        All four adjacency matrices & 6.32 $\pm$ 0.11 & 8.18 $\pm$ 0.14 \\
        \midrule
        PDE-based estimate \citep{liu_real-time_2009} & 15.04 & - \\
        \bottomrule
    \end{tabular}
    \label{tab:g2g}
    \end{center}
    \vspace{-3pt}
    \footnotesize{Values shown are mean abs. error $\pm$ one std. dev. over five random seeds.}
    \vspace{-12pt}
\end{table}

Figure \ref{fig:g2g} shows an example of how our four-edge-dimensional, four-headed NN predicts on the target network.
Interestingly, the NN qualitatively follows the right general trend, but makes a few very large errors in understanding the queue near cycle 10 and near the end of the timeframe.
Note the scale on the plot: a queue length of 60 is not typical on the grid network.
Reexamining Figure \ref{fig:rand}, we see that the random network's irregularity leads to fewer alternate routes than the grid-based one, and therefore perhaps a greater likelihood of bottlenecking.
This means that the queue length of nearly 60 vehicles is atypical for the grid-based one but may be not so out of the ordinary for the random one.
Thus, the underestimation may be due to the fact that the NN had \emph{never seen} that degree of bottlenecking in the grid network.

Table \ref{tab:g2g} shows mean absolute errors for the four-dimensional NN, the NN with no adjoining lane information, as well as the PDE-based baseline.
We see that both the NNs and the PDE-based estimate have much larger errors.
The fact that the baseline error increases suggests that some of the lost accuracy for the NNs may be intrinsic to this network's dynamics presenting a more challenging problem in general.
However, of interest is that the local-lane-information-only NN performs better on both prediction tasks.
Since by construction the lane-local NN does not see what is happening on other lanes, we believe the fact that it now outperforms the adjacency-matrix-using network supports the hypothesis that the different road topology induced different inter-lane and inter-road dynamics as a source of increased error.

Put together, this interpretation justifies the idea of cross-network training to learn dynamics models that may only emerge on particular topologies.
Extending our experiments to more networks, and more complex networks, in this manner is an item of ongoing work.

\section{Conclusion and Future Work}
\label{sec:conclusion}
We introduced the idea of applying attentional operations for deep learning models on lane graphs.
Our results indicate that these particular architectures can be useful on the lane level, where first-principles models are difficult to derive.

We propose two immediate avenues for future work.
First is the application of these NN architectures to more complex and varied traffic scenarios, such as more variable demands, more variable vehicle behaviors, or further applications of the network transfer problem.
The eventual goal of this avenue of work would be the transfer of these NN models from simulation to real-world lane graphs.

The other important avenue is further NN architecture engineering.
We introduced a multi-edge-type attentional layer that can integrate information from different node-to-node relationships in different ways.
Our studies on flattening out the edge types and the comparison with the GCN layer (sections  \ref{sec:flattening} and \ref{subsec:gcn}, respectively) give evidence that this topological information is crucial to learn the complex dynamics on the lane graph.
A shortcoming is that we do not (yet) include the time relationships in these edge types.
Recent items in the NN literature like time convolutional networks \citep{bai_empirical_2018}, and the integration of those with attention \citep{yu_qanet_2018} have challenged the assumption that RNNs are the optimal NN approach for timeseries data (learned downsampling convolutions in the time dimension would also be a more satisfying method of downsampling than our hard 30-second averaging).
An open question is whether the time dimension can be treated as just another edge dimension, or whether it would be more effective to structure the architecture so that time means something distinct from spatial relations.
A deep attention-based model for both space and time is particularly appealing for nonlinear dynamics problems, where the scalar attention scores \eqref{eq:attn_scores} could be used to indicate the importance of different elements of the state space through entire dynamics trajectories according to the learned dynamics model. 
\appendix
\label{app:nn_param_counts}
In our NNs, we set the output dimension (a.k.a. number of units) of all fully-connected layers, as well as the GRU dimensions, to 128 (except for the GCN NN, which had fully-connected layers of dimension 256).
The base graph encoder layer dimension $F$ was 96. Parameter counts for all our NN variants are shown in Table \ref{tab:paramcounts}.
\begin{table}[ht]
    \centering
    \caption{Parameter counts for NN configurations}
    \label{tab:paramcounts}
    \begin{tabulary}{\columnwidth}{l l R}
        \toprule
        From: & NN Configuration & Param. count \\
        \midrule
        \multirow{4}{*}{Table \ref{tab:results}} & $A=I$& 385,027 \\
        & +$A_\textnormal{downstream}$, $\{I, A_\textnormal{upstream}\}$, $\{I, A_\textnormal{neighbors}\}$ & 486,403 \\
        & +$A_\textnormal{upstream}$ & 587,779 \\
        & +$A_\textnormal{neighbors}$ & 689,155 \\
        \midrule
        \multirow{4}{*}{Table \ref{tab:ablation}} & approx. $\sfrac{1}{4}$-size attn. dim. ($F=32$) & 384,003 \\
        & Flattened adjacency matrix & 385,027 \\
        & One (4x-size) attention head & 689,155 \\
        & GCN (all versions) & 481,027 \\
        \bottomrule
    \end{tabulary}
\end{table} \vspace{-12pt}

\section*{Acknowledgements}
This research was supported by the National Science Foundation under grant CNS-1545116 and by Berkeley DeepDrive.
We also made use of the Savio computational cluster provided by the Berkeley Research Computing program at the University of California, Berkeley.

We thank Simon Sohrt of the Institute of Transport and Automation Technology at Leibniz Universit\"{a}t Hannover and three anonymous reviewers for their reading and feedback.

\bibliographystyle{IEEEtran}

\end{document}